\documentclass[3p,times,procedia]{elsarticle}
\usepackage{ecrc}
\usepackage[bookmarks=false]{hyperref}
    \hypersetup{colorlinks,
      linkcolor=blue,
      citecolor=blue,
      urlcolor=blue}
\usepackage{graphicx}
\usepackage{subfig}
\usepackage{amssymb,amsmath}
\usepackage{ecrc}
\usepackage[capitalise]{cleveref}
\usepackage{multirow}
\volume{00}

\firstpage{1}

\journalname{Procedia Computer Science}

\runauth{Anna~V.~Bubnova}
\runtitle{Approach of variable clustering and compression for learning large Bayesian networks}

\jid{procs}

\usepackage{amssymb}
\usepackage[figuresright]{rotating}
\begin{document}
\begin{frontmatter}



%%\dochead{ 11th International Young Scientist Conference on Computational Science}%
%% e.g. \dochead{17th International Conference on Dynamical Processes in Excited States of Solids}

\title{Approach of variable clustering and compression for learning large Bayesian networks}


\author[a]{Anna~V.~Bubnova\corref{cor1}} 

 \address[a]{ITMO University, Saint-Petersburg, Russia}

\begin{abstract}
This paper describes a new approach for learning structures of large Bayesian networks based on blocks resulting from feature space clustering. This clustering is obtained using normalized mutual information. And the subsequent aggregation of blocks is done using classical learning methods except that they are input with compressed information about combinations of feature values for each block. Validation of this approach is done for Hill-Climbing as a graph enumeration algorithm for two score functions: BIC and MI. In this way, potentially parallelizable block learning can be implemented even for those score functions that are considered unsuitable for parallelizable learning. The advantage of the approach is evaluated in terms of speed of work as well as the accuracy of the found structures.
\end{abstract}

\cortext[cor1]{- corresponding author (anna.v.bubnova@gmail.com)}
\end{frontmatter}

% \linenumbers


%\enlargethispage{-7mm}
\section{Introduction}
In recent years, the amount of statistical data available has increased many times over, not only in terms of the number of observations, but also in terms of the parameters of the objects under study too. One of the universal tools for representing objects of random nature with complex systems of characteristic dependencies is Bayesian networks. They are suitable for a wide class of tasks: modelling, prediction, gap recovery, outlier search, etc. However, for data with a large number of characteristics, finding a suitable Bayesian network in the sense of a directed acyclic graph structure (DAG) that describes the dependence of the features is difficult because the number of possible DAGs grows super exponentially with the number of nodes \cite{robinson1973counting}. There are many \cite{scutari2019learns} approaches to solve this problem, the most effective of which include paralleling learning approaches at the node level \cite{ramsey2017million}. On the other hand, even in this case the number of potential parent combinations for each node is exponential to the number of nodes when considered separately. It leaves the opportunity for further acceleration by reducing the dimensionality of this space even for situations where node-level paralleling is not possible, e.g. for MI-based learning \cite{chickering2002optimal}. 

There are works based on this idea in which parents are searched locally, e.g. for ``nearest" nodes found using kNN method and some divergence for pairs of parameters \cite{liu2016inference}. The main difficulty of such approaches occurs at the point where the local structures need to be collected together, since edges in local structures can form loops in aggregate, and there is a need to remove redundant edges that may have resulted from the localization of the search space. Another difficulty is that the mutual information used is not normalized and may give priority to pairs of variables with a large number of values. It is possible to solve these problems by taking normalized divergence and considering clusters instead of kNN. This idea is implemented in the paper \cite{njah2019deep}, however, only for Hierarchical Bayesian Networks.

Continuing with this idea, this paper takes the same normalized mutual information and considers independent clusters. Of course, these clusters could be connected in a consistent way using already known algorithms, for example, Hill-Climbing. However, with such a connection we just set some initial graph, but we do not reduce the search space. To do this, we must somehow reduce the number of nodes representing the characteristics of the object. In this paper, this is done by compressing combinations of values for each cluster of features, in one case by reducing to the most frequent combinations, in another by clustering based on Hamming Distance. And unlike the work \cite{njah2019deep} already mentioned, since a clique can be anywhere in the structure, not just at the lowest levels, this compressed information, represented as a node, can both receive information from other nodes and give it away.

The need for compression with loss is dictated by the fact that the number of features in a cluster, in the most ideal case in terms of search space size, grows as the total number of features increases, and the number of value combinations grows exponentially with the size of the cluster. It is exceptionally rare to find datasets where the number of observations is so large as to sufficiently cover every combination of values. A more detailed discussion of the optimal number of clusters and the size of clusters can be found in \cref{background}.

The above compression algorithms, unfortunately, cannot be easily replaced by classical alphabetic encryption approaches since in the case of cluster ``letters" are not repeated. And there is also a difficulty with lossy compression algorithms since they are mostly adapted to deal with continuous values. But in this paper, even in the case of some continuous parameters structure learning is done for discretized data. And hence compression to a discrete value is required. Although it would be interesting to investigate this option in the future.

Thus the approach presented here solves these problems, demonstrates its validity in the sense of speeding up the basic algorithm without significant loss of accuracy. Just as it can be easily extended by new methods of combination compression, adding another graph structure enumeration algorithm like GES and introducing new score functions. The approach does not impose any restrictions on this, although new variants require further investigation.

\begin{figure}[ht!]\vspace*{4pt} 
\centerline{\includegraphics[trim=10 510 125 0,clip,
width=0.6\textwidth]{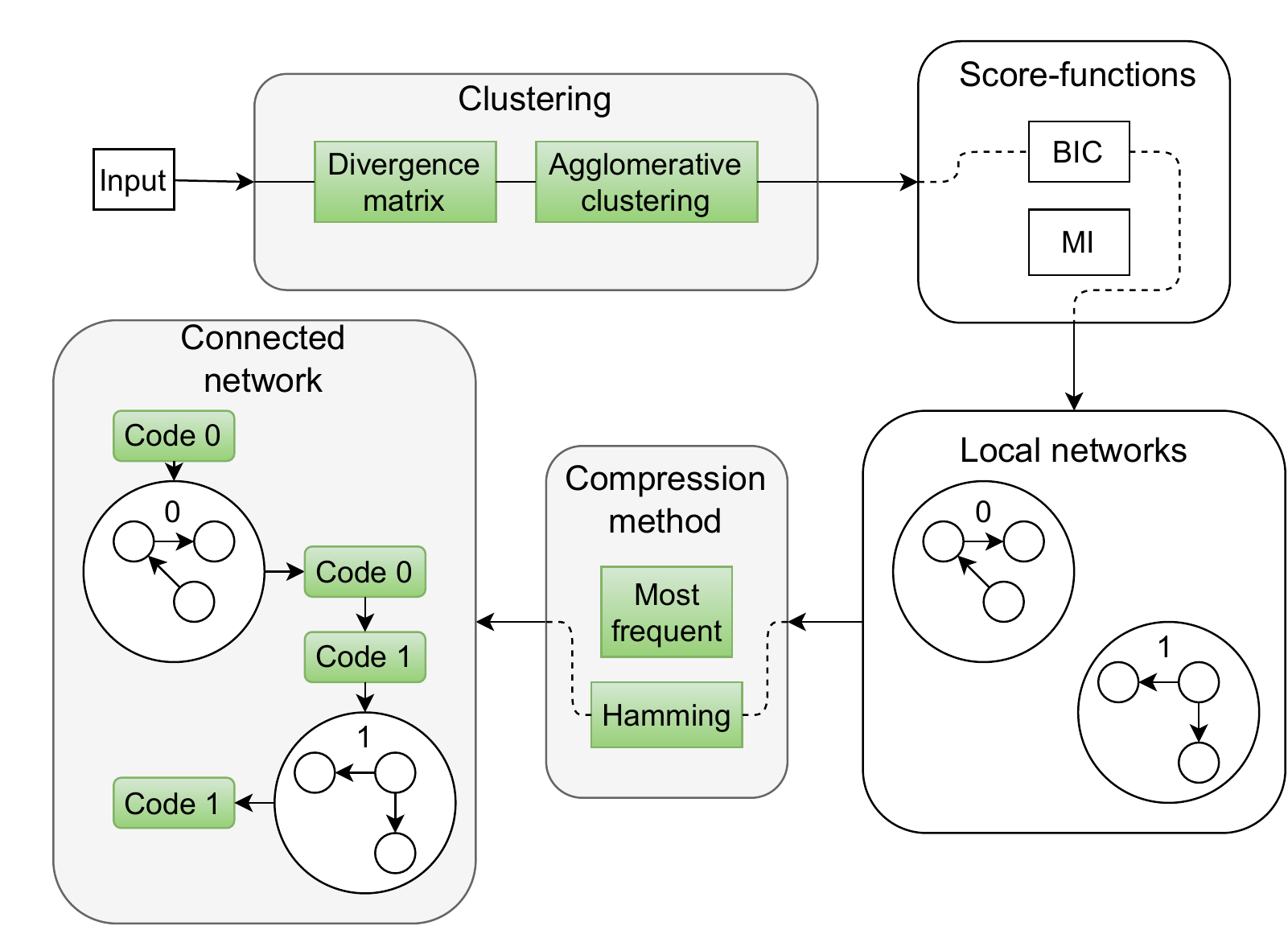}}
\caption{Scheme of the approach implementation. The dotted line shows examples of option choices. The main elements of the proposed approach are highlighted in green.} \label{fig:pipeline}

\end{figure}

A scheme of the proposed approach is shown in \cref{fig:pipeline}. 

\begin{nomenclature}
\begin{deflist}[WMO]
\defitem{BN}\defterm{Bayesian network}
\defitem{HC}\defterm{Hill-Climbing}
\defitem{DAG}\defterm{Directed acyclic graph}
\defitem{MI}\defterm{Mutual information}
\defitem{BIC}\defterm{Bayesian information criterion}
\end{deflist}
\end{nomenclature}

\section{Related Work}\label{Related}
This section will present brief information and references to works on structural learning algorithms as well as information compression approaches.

The closest to this is the work \cite{njah2019deep} on Hierarchical Bayesian Networks, which uses the same information divergence to determine the clique nodes. Then these are hung on additional nodes of compressed values obtained by the method of most frequent combinations and by recursive repetition of this procedure form a hierarchical Bayesian network. 

Authors of the paper \cite{scutari2019learns} compare many algorithms for structure search. There is no point in comparing as the speed is expected to depend on the base algorithm, but it is interesting which algorithms from this work, combined with the proposed approach will give a speedup.

Parallelisation, such as in fGES \cite{ramsey2017million}, is a good way to accelerate. However, the functions must have local consistency \cite{chickering2002optimal} for this algorithm. For example, MI and K2 do not have this property. The proposed approach does not impose such restrictions but also allows for parallelisation.

\section{Backgrounds}\label{background}
In general, Bayesian networks are a way to represent a factorization of probability distribution $P(X_1,\dots,X_p)$ using a Directed Acyclic Graph (DAG) $G = (V, E)$, where $V=\{X_j\}_{j=1}^p$. By factorization in this case we mean the decomposition of joint probabilities into the product of conditional probabilities. The conditions are the sets of parents $\Pi_X=\{X_k:X_k\neq X, (X_k,X)\in E\}$ generated by the structure of graph $G$:
\begin{equation}
    P(X_1,\dots,X_p )=\prod_{j=1}^p P(X_j |\Pi_{X_j}).
\end{equation}
This paper considers only the group of algorithms for finding $G$ from the available data $D$, which best describe the factorization of the observed joint distribution relying on score functions. From a formal point of view, such algorithms walk in DAG space and look for the maximum or minimum sum of some local score function $s$:
\begin{equation}
    S(G, D)=\sum_{j=1}^p s(X_j|\Pi_{X_j}) \rightarrow \max(\min).
\end{equation}

The aim of this work is to implement and compare the quality and speed of algorithms that look for optimal general and local structures for clusters in terms of MI and BIC.

Clustering is done by the following divergence between 0 and 1:
\begin{equation}
d(X,Y) = 1 - NMI(X,Y),
\end{equation}
where $NMI(X,Y)$ is the normalized mutual information that it may also be known as Symmetric Uncertainty. For a pair of random variables $X$ and $Y$, it is calculated through MI and entropy (E) as follows:
\begin{equation}
NMI(X,Y)=\frac{2\cdot MI(X,Y)}{E(X)+E(Y)}.
\end{equation}

Unfortunately, we do not know the optimal number of clusters however given the super exponentiality of the DAG space we need the maximum number of clusters and the maximum cluster size to be minimal. In the best case, this will be $\sqrt{p}$ and then the number of combinations in the cluster is on the order of $\exp{\sqrt{p}}$, which leads us to the necessity to compress the combinations.

The first way of compressing values relies on Hamming Distance (HD) clustering, which counts the number of changes that need to be done to one combination to get another. Another method is based on coding the most frequent ones. This procedure can be described as follows. Let a combination $I_i = (x_{1i},\dots, x_{ui})$ have a probability estimated and sorted in descending order:
\begin{equation}
P(I_1) \ge \dots \ge P(I_k).
\end{equation}

For a given $\alpha$ the first $l$ combinations are searched such that:

\begin{equation}
\sum_{i=1}^{l-1} P(I_i) < 1 - \alpha, \sum_{i=1}^{l} P(I_i) \ge 1 - \alpha.
\end{equation}

And $g$ code for $I_i$ is given according to the following principle:

\begin{equation}
g(I_i) =
 \begin{cases}
   i-1, &\text{ $i \le l$}\\
   g\left(\arg\min_{j\le i} HD(I_j, I_i)\right), &\text{ $i>l$}
 \end{cases}
\end{equation}

\section{Algorithms and methods}
\subsection{General approach description}
The approach is based on the following steps:
\begin{enumerate}
\item Calculation of pairwise divergences between features and clustering based on them;
\item Local structure learning for each cluster;
\item Compress combinations for parameters in each cluster and replace with code;
\item Learning the global structure on data replaced by codes;
\item Connecting local networks via support code nodes according to the global structure for them.

\end{enumerate}

\subsection{Clustering}
The first step before structural learning is the calculation of pairwise divergences for features and agglomerative clustering. We do not know the optimal number of clusters, so the algorithm is considered for different threshold values on a uniform grid from 0 to 1. However, there is also the recommendation for this threshold, which boils down to minimizing the maximum number of clusters and sizes of clusters. It will be shown that this recommendation allows the algorithm consistently to run faster than the average for a random threshold.

\subsection{Compression}

For compression using the most frequent, a level of 95 per cent is chosen, which must cover the most frequent combinations. There is also a requirement to have at least 5 observations for each the most frequent combination. The least frequent combinations are matched to the closest frequent combination in terms of Hamming Distance, and in the case of equal distances to several frequent combinations, to the most frequent of them. Then each most frequent combination is given its own number or symbol that acts as input to the global structure search algorithm instead of the values in the observation. The previously introduced constraints may lead to the fact that this compression will not be possible for all clustering.

A threshold of 95 per cent in agglomerative clustering is chosen for Hamming Distance compression. There are no other restrictions, which makes this compression possible for any feature clustering. In this case each cluster is given its own number or symbol that gets input to the global structure search algorithm instead of the values in the observation. 

These kinds of constraints are needed to simplify the evaluation of the algorithm's performance, and of course more research is needed on the effect of thresholds on the results of the approach.

\subsection{Connecting}
\begin{figure}[ht!]\vspace*{4pt} 
\centerline{\includegraphics[trim=10 600 165 0,clip,
width=0.6\textwidth]{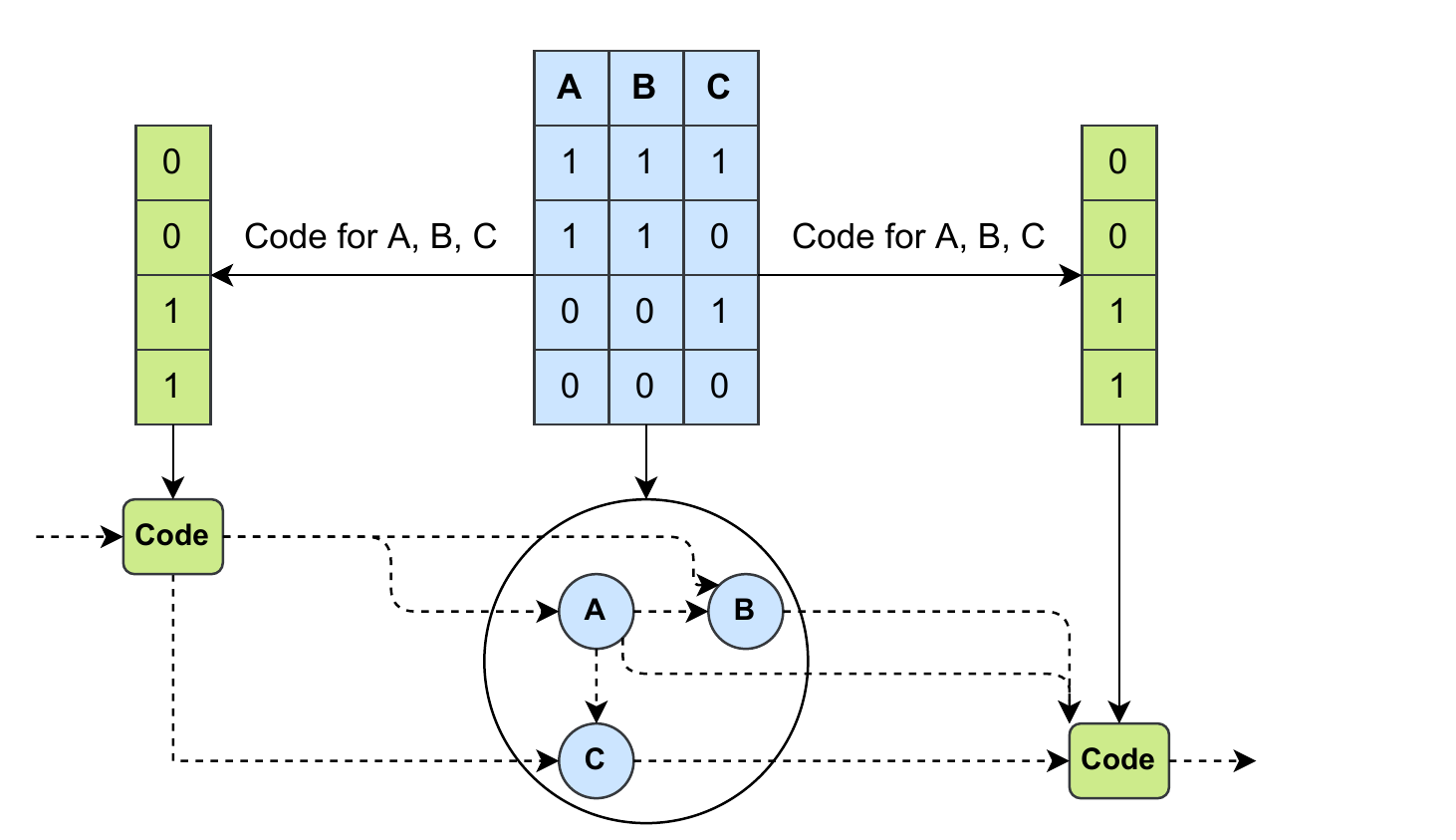}}
\caption{Scheme for connecting code nodes and cluster nodes. The dotted arrows show the links within the global network.} \label{fig:coding}

\end{figure}

The local structures are not connected directly. For each local structure, two identical support nodes are created that correspond to the code for that cluster. One is set at the top level and from it the edges lead to all nodes in the cluster, the other node is set at the bottom level and the edges lead to it from all nodes in the cluster. \cref{fig:coding} describes the scheme of this connection. Considering that compression is lossy, this configuration may lead to distortion of the simulation distribution in the cluster or degradation the quality of gaps recovery.

\section{Experiments and Results}
\subsection{Datasets Description}
The approach is compared on four synthetic datasets. Datasets SANGIOVESE, MEHRA, HEPAR II, and PATHFINDER represent the synthetic data sampled from $bnlearn$ networks \cite{scutari2020package}. The last two have only discrete values and the others are mixed. For more information about datasets, see \cref{table_datasets'_info}. Purely continuous synthetic data sets are not considered here, as they are usually represented by multivariate Gaussian distributions with values clustered around a single center. Such data would be difficult to compress in terms of clustering and frequency of values combinations. The advantage of synthetic data is that its network is known, allowing structures to be comparable. 

\begin{table}[!ht]
\caption{Number of nodes and observations for each dataset.}
\label{table_datasets'_info}
\begin{center}
\begin{tabular}{|c|c|c|c|c|c|}
\hline
Parameters & Sangiovese & Mehra & Hepar II & Pathfinder \\  
\hline
Total nodes & 15 & 24 & 70 & 109 \\
\hline
Discrete  nodes & 1 & 8 & 70 & 109 \\
\hline
Continuous nodes & 14 & 16 & 0 & 0 \\
\hline
Size & 4500 & 4500 & 3000 & 3000  \\
\hline
\end{tabular}
\end {center}
\end{table}

The information divergence used in the considered implementation of the approach works exclusively with discrete data. Due to this, the values of each continuous feature were divided into five bins with approximately the same frequency, and the data were replaced by the labels of these bins.
    
\subsection{Comparison of Bayesian network learning algorithms} \label{experiments} 
\subsubsection{Comparisons of time and SHD for synthetic data}
\begin{figure}[!ht]
\centering
    \subfloat[]{\label{geo}\includegraphics[width=0.49\textwidth]{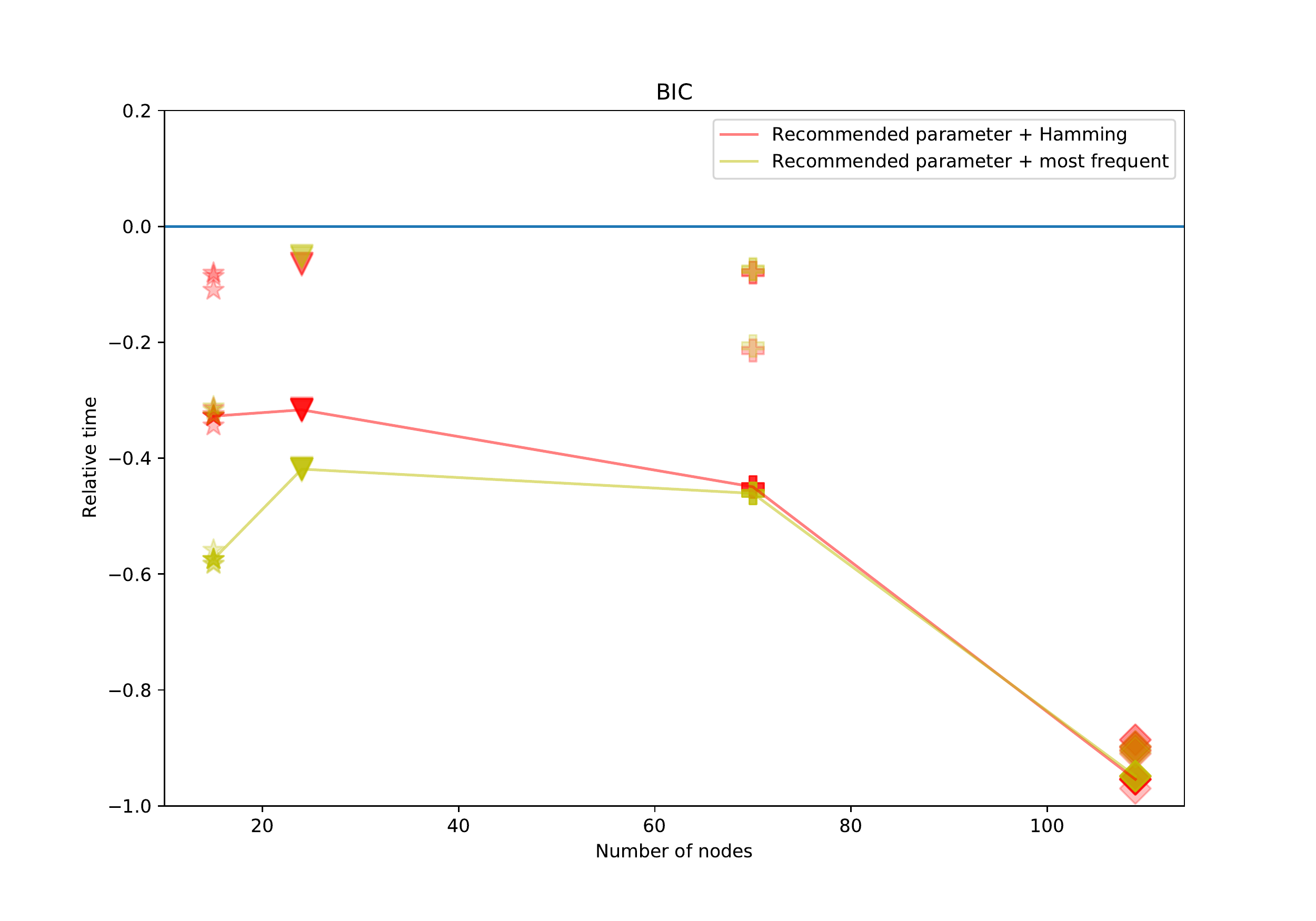}} 
    \subfloat[]{\label{geo}\includegraphics[width=0.49\textwidth]{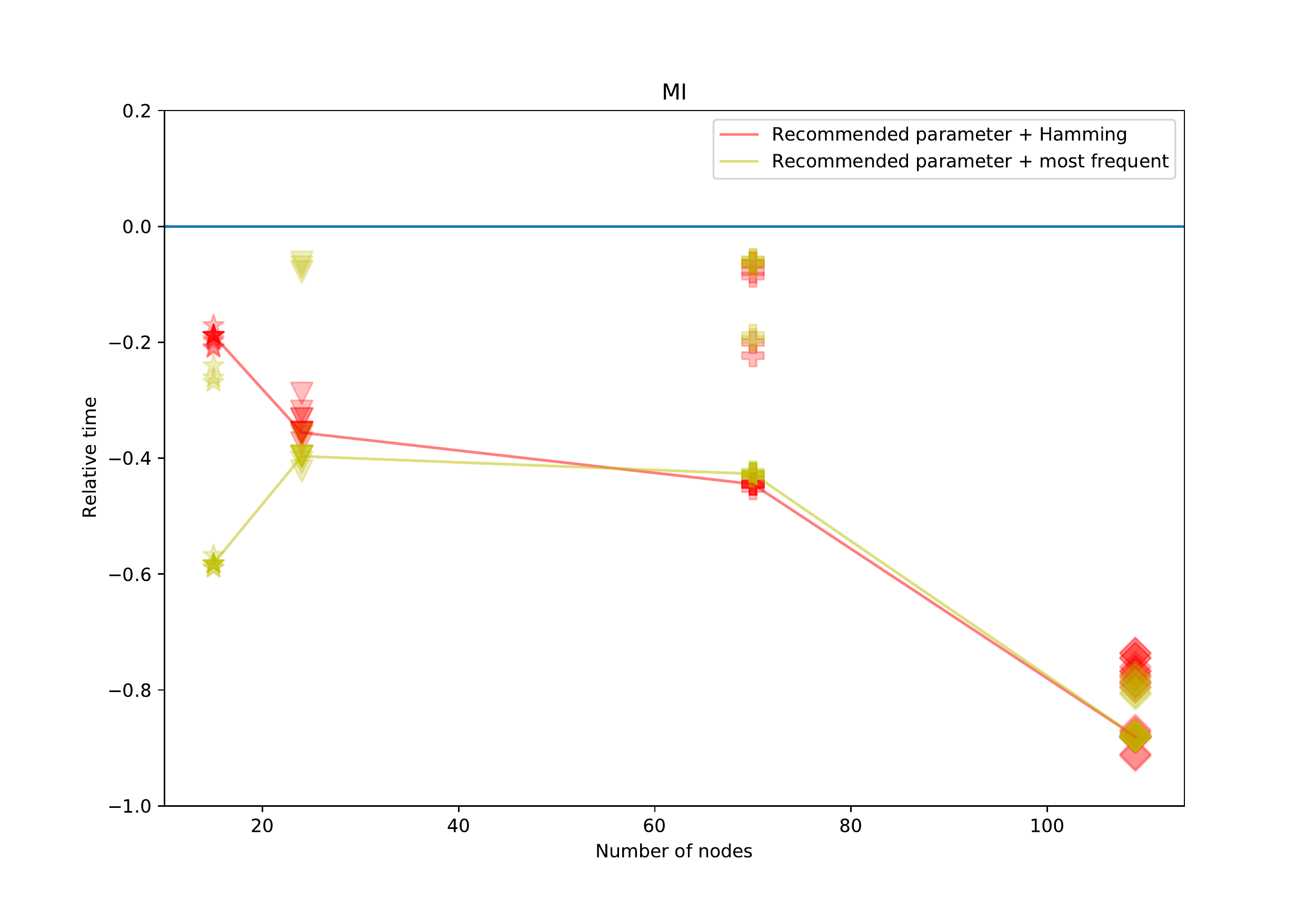}} 
   
    \caption{The ratio of time normalized to 1 for the proposed approach is relative to the classical algorithm for BIC and MI score functions.}
    \label{fig:synth_time}
\end{figure}

The goal of the proposed approach is to reduce the running time of the algorithm for searching the structure of the Bayesian network with a possibly insignificant loss in the quality. There is also an assumption that the variant minimizing the search space should have the highest speed. Therefore in the first experiment the running times of the classical algorithm and the proposed algorithm for the cluster splitting thresholds grid with a step of 0.1 and two proposed compression options were compared. All measurements are performed on the same PC (i5-9300H CPU, 8GB RAM). The quality measure is Structural Hamming Distance (SHD), which for a pair of graphs shows how many deletions, additions and inversions of edges it takes to get a second graph from one. Since we are comparing with the classical algorithm for both time and SHD, the ratio of estimation for the approach to estimation for the unmodified algorithm minus one is calculated. The resulting values should be less than zero and ideally close to -1. \cref{fig:synth_time} shows the time comparison results for the score functions BIC and MI. Each point corresponds to the running of the modified algorithm with some clustering. The x-axis displays the number of nodes in the network, i.e. the number of parameters in the dataset. It can be seen that the proposed approach has a lower growth rate than the classical approach, and the recommended clustering parameter does give the fastest option.

\begin{figure}[!ht]
\centering
    
    \subfloat[]{\label{socio}\includegraphics[width=0.49\textwidth]{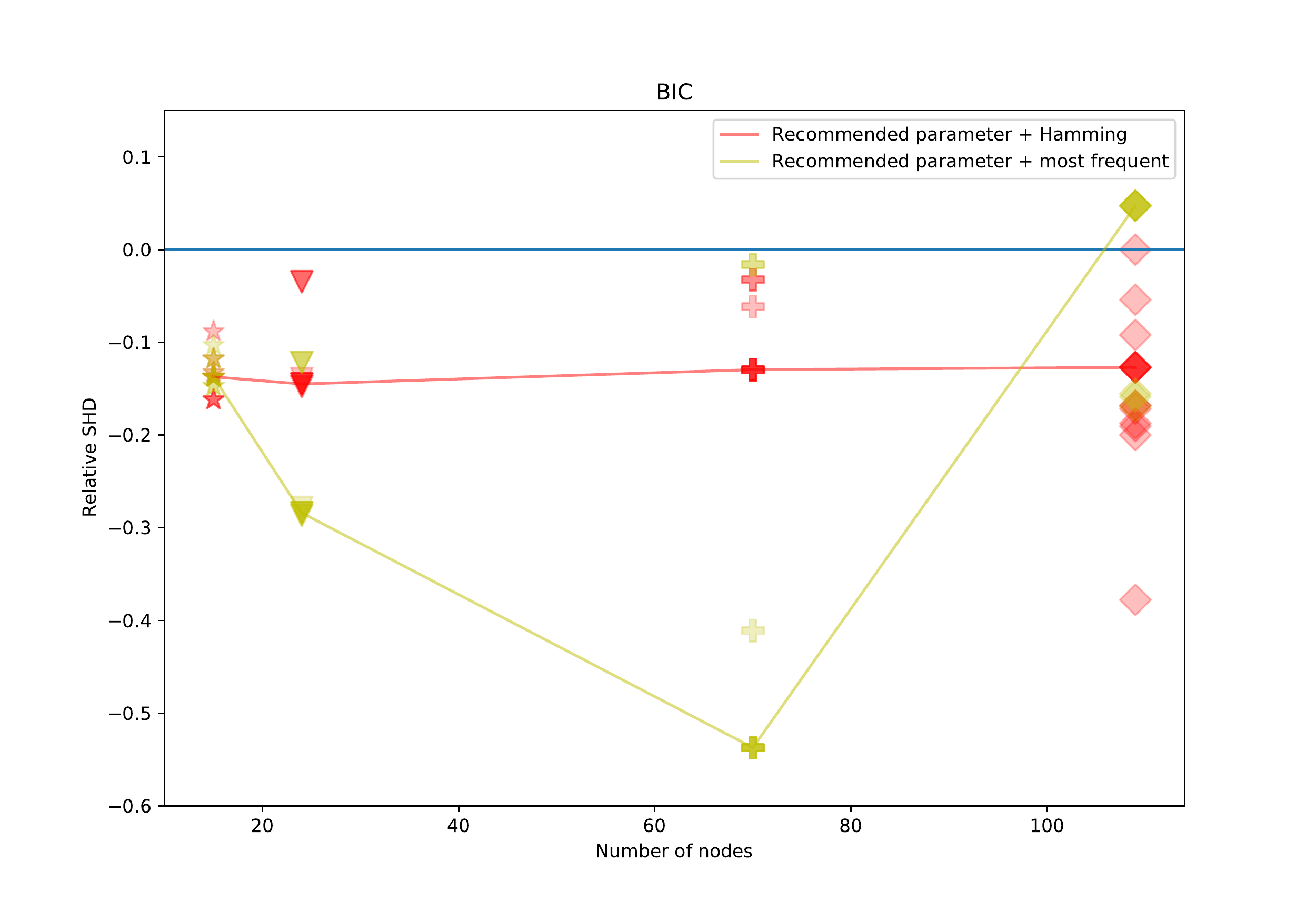}}
    \subfloat[]{\label{socio}\includegraphics[width=0.49\textwidth]{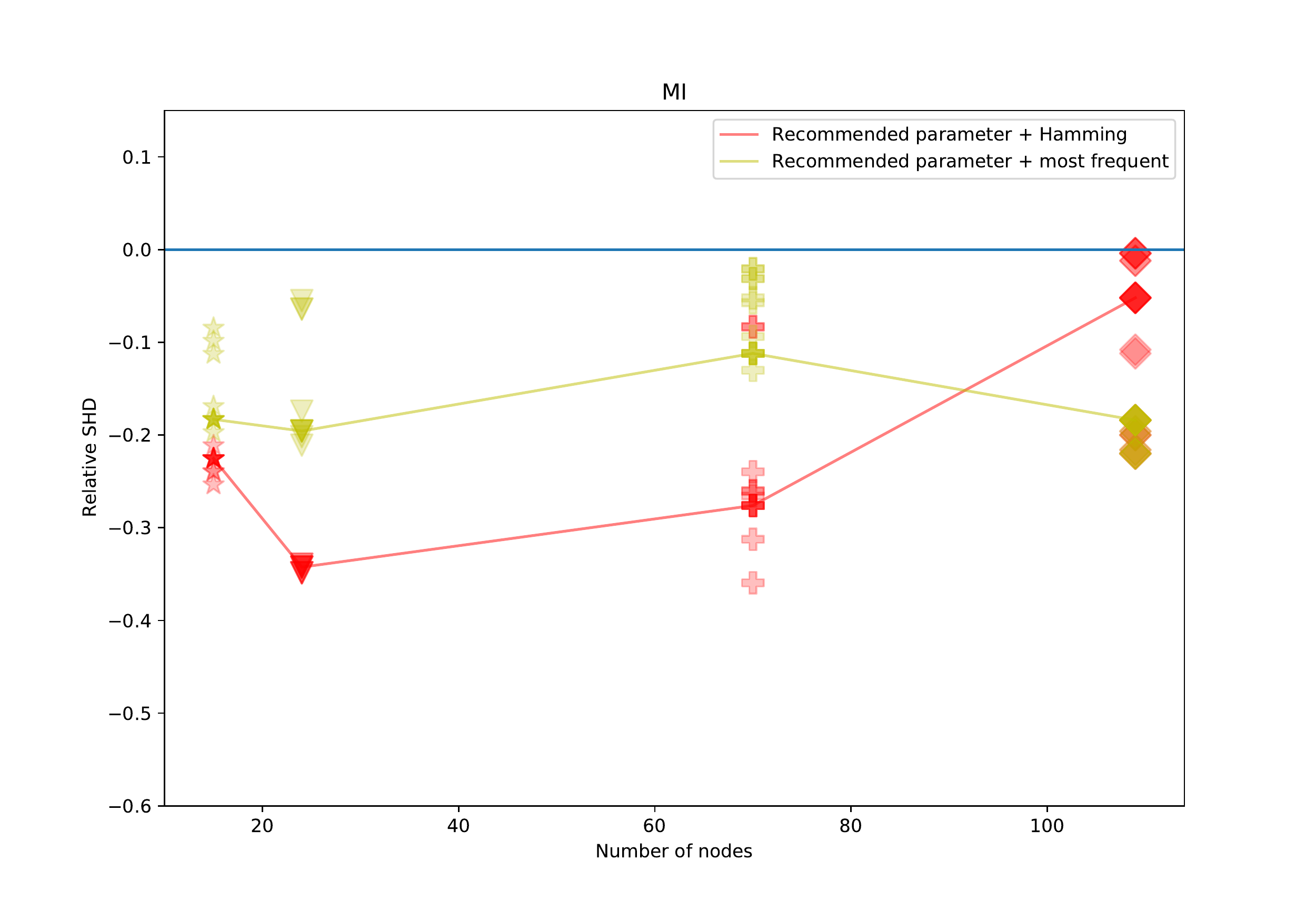}}
    \caption{The ratio of SHD normalized to 1 for the proposed approach is relative to the classical algorithm for BIC and MI score functions.}
    \label{fig:synth_shd}
\end{figure}

\cref{fig:synth_shd} shows the SHD comparison results for the score functions BIC and MI. Each point corresponds to the running of the modified algorithm with some clustering. The x-axis displays the number of nodes in the network, i.e. the number of parameters in the dataset. It can be seen that the proposed clustering parameter does not always give the best option in terms of quality. However the clustering for almost all implementations gave a closer result to the original network than the classical one. In the future it is planned to consider recursive clustering for large networks such as PATHFINDER, since the local clusters are already large enough, and the drop in quality may be related to this fact.

\subsubsection{Comparisons of accuracy for synthetic data}
A separate question is whether it is necessary to connect local clusters using compressed combination variables or whether cluster-based local networks are sufficient. On the one hand, if the analysis of the divergence matrix shows there are nodes not included in any cluster, the nodes with codes will affect the network structure for them. On the other hand, this is not a frequent situation, and the fastest implementation requires cluster size averaging. So compression effects may arise in the modeling and gap-filling problem since it is only through them that additional information can be transferred from one cluster to another.

For this reason, this experiment investigated the ability of connected nodes with codes and separated networks to recover skips. For each instance, the value of each parameter is removed for the separated subnets and for the connected ones, the code belonging to the cluster containing that parameter is additionally removed. Then the value is recovered from the networks, and the recovery accuracy is calculated for all samples. The ratio of the estimate for the connected networks to the estimate for the unconnected networks is also calculated and normalized to one. In the case of accuracy, a ratio greater zero shows higher model quality. 

\begin{figure}[!ht]
\centering
    \subfloat[]{\label{geo}\includegraphics[width=0.49\textwidth]{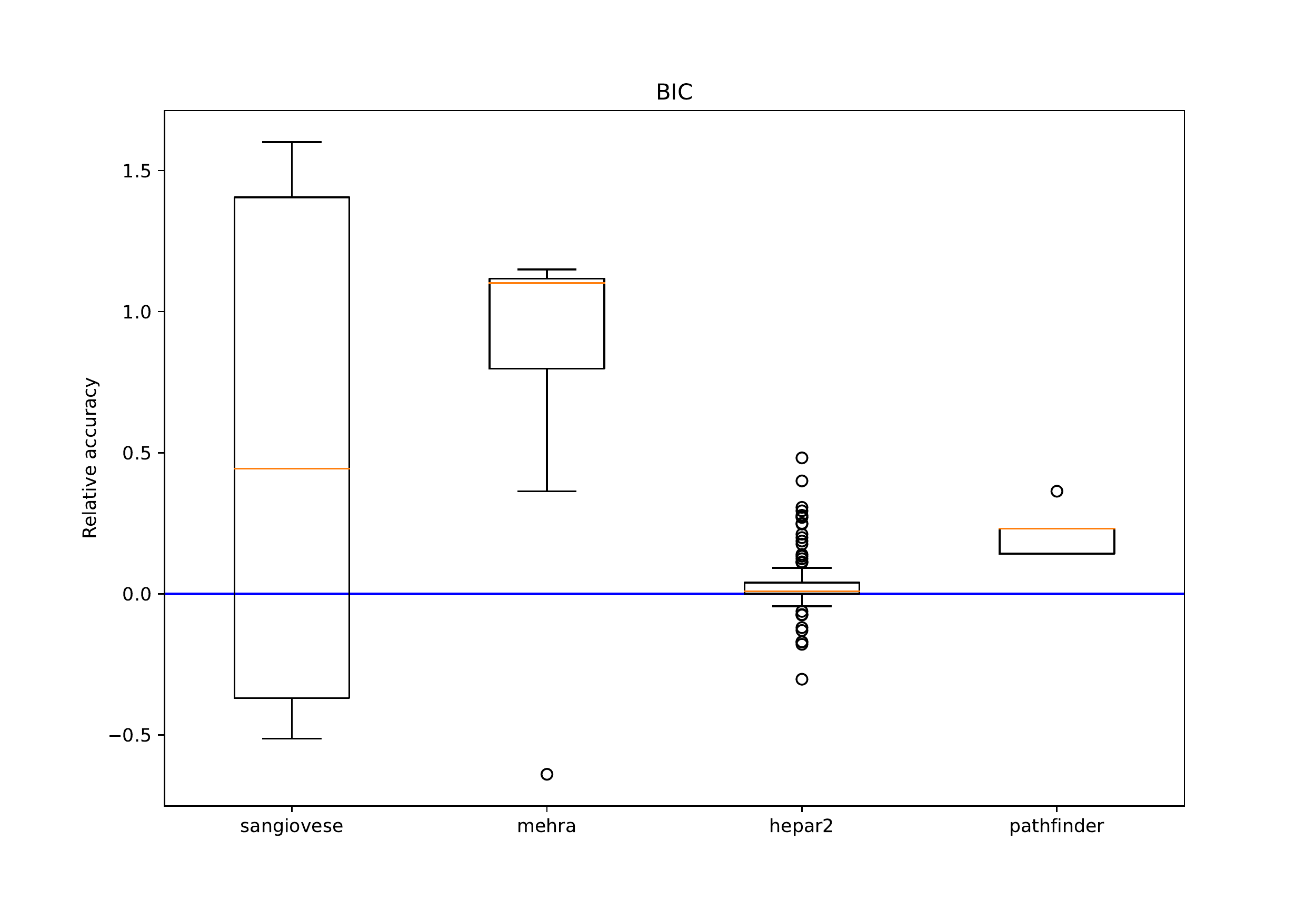}} 
    \subfloat[]{\label{socio}\includegraphics[width=0.49\textwidth]{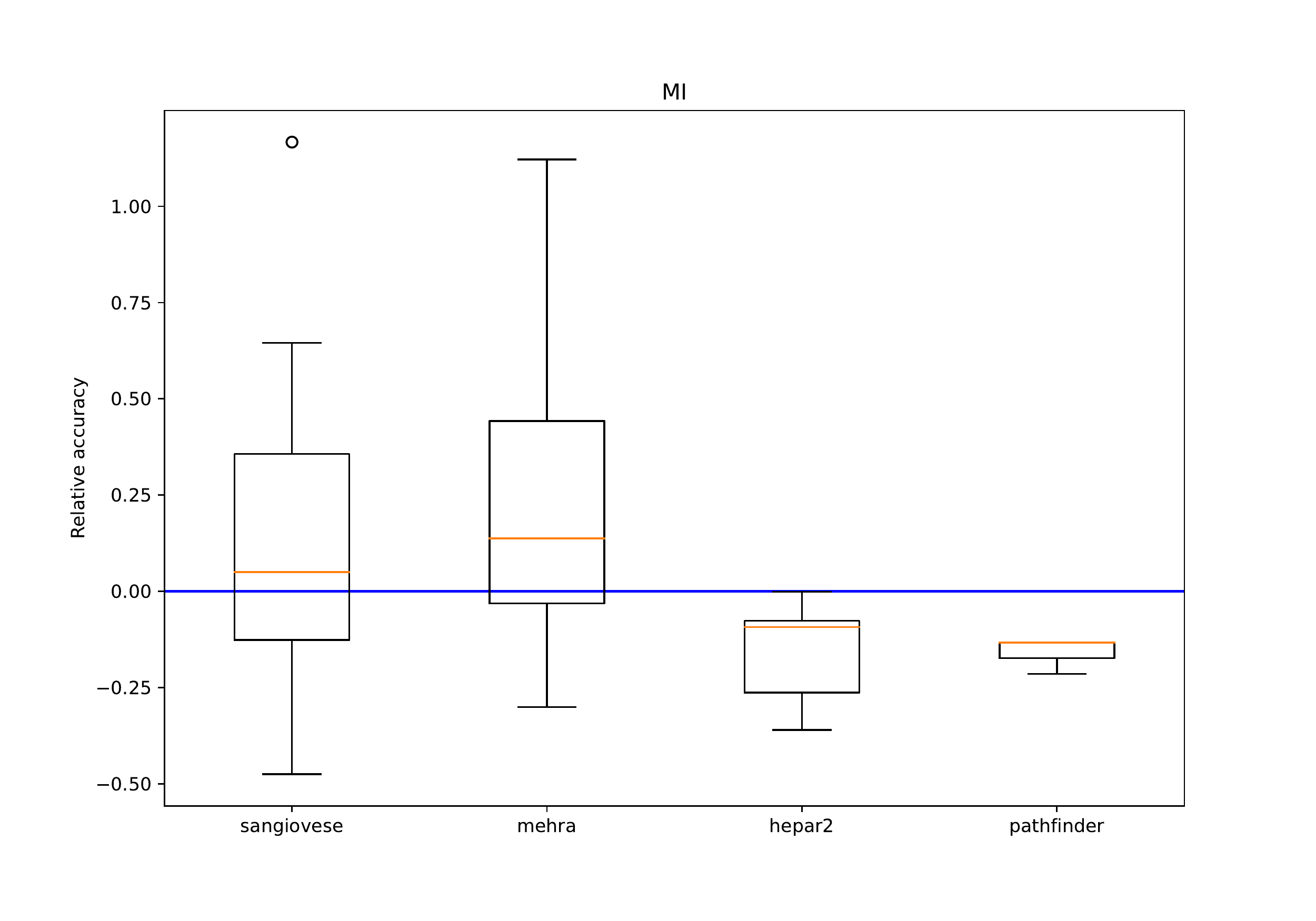}}
    \caption{The ratio of accuracy normalized to 1 for the proposed approach is relative to the classical algorithm for clusters without compression data.}
    \label{fig:synth_acc}
\end{figure}

\cref{fig:synth_acc} shows that if the model for connected networks shows a different result, it is mostly positive and quite noticeable for medium-sized networks. The drop in quality for larger networks and especially for MI may be because the number of combinations and codes increases as the cluster size increases. And with a limited number of observations, the coverage of these codes drops. MI's distribution parameters, on the other hand, are closely related to the number of value combinations \cite{de2006scoring}, which may also cause local networks to be combined more incorrectly than BIC, for example. In the future, further research is planned for datasets with large numbers of observations for other compression methods and basic algorithms.

\section{Conclusion}
This paper presents an approach for training a Bayesian network with clustering based on information divergence of features and lossy compression for combinations of values in clusters. Experiments have confirmed a significant acceleration of Hill-Climbing when modified using this approach, as well as higher accuracy of the model built with compressed information in the recovery problem in most cases, especially for medium-sized networks.

In the future, it is planned to implement parallelization of this approach, to expand the pool of score functions with LL, AIC, and model-based and base algorithms, e.g., with GES. It would be interesting to consider Rajski Distance and other distances between variables for clustering. There are also plans to investigate distances for mixed types of values for data compression, such as Gower's Distance and metrics for continuous, to avoid discretization. Studying how hyper-parameters of compression and quality of combination coverage affect modeling and prediction for connected networks is another area for development.

All source code and materials used in the paper are available in the repository \citep{github-sources}.

%% References

%The citation must be used in following style: \cite{article-minimal} \cite{article-full} \cite{article-crossref} \cite{whole-journal}.

\bibliography{mybibliography}
\bibliographystyle{elsarticle-harv}




\clearpage

%%%% This page is for instructions only, once the article is finalize please omit the below text before creating the final PDF

\end{document}